\def\year{2020}\relax
\documentclass[letterpaper]{article} 
\usepackage{aaai20}  
\usepackage{times}  
\usepackage{helvet} 
\usepackage{courier}  
\usepackage[hyphens]{url}  
\usepackage{graphicx} 
\usepackage{comment}
\usepackage{graphicx}  
\usepackage{multirow}
\usepackage{xcolor}
\usepackage{lscape}

\title{Assessing Fairness in Classification Parity of Machine Learning Models in Healthcare}
\author{\Large \textbf{Ming Yuan,\textsuperscript{2}
Vikas Kumar,\textsuperscript{1}
Muhammad Aurangzeb Ahmad,\textsuperscript{1,3}
Ankur Teredesai,\textsuperscript{1,2} }\\
\textsuperscript{1}{KenSci Inc, Seattle, WA}\\
\textsuperscript{2}{Department of Computer Science, University of Washington - Tacoma}\\
\textsuperscript{3}{Department of Computer Science, University of Washington - Bothell}

\\ 
\\ 
mirandayuan09@gmail.com, \{vikas, muhammad, ankur\}@kensci.com \\ 
}

\begin{document}

\maketitle

\begin{abstract}
Fairness in AI and machine learning systems has become a fundamental problem in the accountability of AI
systems. While the need for accountability of AI models
is near ubiquitous, healthcare in particular is a challenging field where accountability of such systems takes
upon additional importance, as decisions in healthcare
can have life altering consequences. In this paper we
present preliminary results on fairness in the context of
classification parity in healthcare. We also present some
exploratory methods to improve fairness and choosing
appropriate classification algorithms in the context of
healthcare.
\end{abstract}

\section{Introduction}
Although machine learning has been around for just over
sixty years, it is only in the last decade or so that its influence on society at large is being felt, as systems powered
by machine learning are now impacting the lives of billions
of people. For instance, recommendation systems that suggest items to people by inferring their preferences play a pivotal role in most e-commerce sites such as Amazon, Netflix,
Alibaba etc. Other example include predicting crimes for active policing, predicting risk of re-offence to facilitate sentencing, financial decision making, and decision making in healthcare. Given that
many applications of machine learning have potential life
changing implications, fairness in machine learning has thus
become a critical issue.

Additionally, the quest for fairness
in machine learning is motivated in many domains by the
desire to adhere to national and international legislation, for example
the GDPR in the European Union, the Universal Declaration of Human Rights \cite{assembly1948universal} in the context of the
digital age \cite{zliobaite2015survey} etc. The quest for fairness in machine learning is part of the
larger enterprise of creating responsible machine learning
systems that engender trust and ensure transparency of
the machine learning methods being used \cite{zliobaite2015survey}.
This involves explainability of the machine learning model
and often requires guarantees regarding what would happen
when the algorithms involved in making decisions that
impact lives of people. Fairness in machine learning is especially critical for minority or vulnerable  are more likely to be affected by decision making by automated by machine learning systems. Thus, creating machine learning
systems that are fair is pivotal to upholding the social contract. While there is wide agreement on the need for fairness in machine learning, there is
no single notion of fairness that can be applied for all use cases. The reason
for this being that fairness can refer to disparate but related
concepts in different contexts.

Though it is universally acknowledged that fairness is
critical in most domains, in certain applications in the
judicial system or in healthcare its importance and impact is paramount. This is because the algorithms used in
the field of healthcare can be both widespread and
specialized, which may require additional constraints to consider to build a fair system \cite{ahmad2018interpretable}. To illustrate how the usage of machine learning models can affect and bias models in critical domains, we focus on healthcare in this paper as of
a domain where fairness in machine learning can have life
changing consequences. While there are multiple notions of fairness in machine learning, we focus on the classification parity with respect to protected features such as age, gender, and race. Additionally we measure and address fairness
for various classification tasks in the performance of Machine Learning algorithms over various 
healthcare datasets. Specifically, we address the following
problems:
\begin{itemize}
    \item Measure fairness as classification parity in the context of
predictive performance of machine learning methods
    \item Determine how the ablation of protected features affects
the performance and fairness of machine learning methods in general and also in via sampling
    \item Determine fairness threshold i.e., a threshold for machine
learning models where the models are relatively fair and
as wells as predictive performance is sufficiently good
\end{itemize}
We use three publicly available datasets to explore the
question of fairness outlined here. The healthcare datasets that are
used are somewhat limited in terms of the small size of
the datasets. Because of this limitation the differences between the predictive performance of certain prediction models with different classification thresholds may not be statistically significant. In this paper, however, our goal is to show
the feasibility of the techniques employed. We do plan to
address limitation of publicly available dataset in the future
by deploying the framework outlined in this paper in a large
hospital system in the mid-West in a real world clinical setting.
 
\section{Related Work}
Bias is inevitable in any sufficiently complex dataset. The data collection and capture process tends to capture only aspects of the phenomenon of
interest, and hidden assumptions may lie in data collection,
processing and analysis. It is thus unavoidable that when
machine learning algorithms learn from data, intentional or
unintentional discrimination can result \cite{barocas2016big}. There is extensive literature on issues related to
fairness in different application domains within machine
learning e.g., natural language processing, image classification, target advertising, and judicial sentencing. Studies from
these various domains prove that data bias is
inherit in many disciplines which in turn creates disparate
treatment effects across categories. To address these limitations, many commercial organizations have released
software to address fairness in machine learning models e.g.,
Google, Amazon , IBM etc. More recent developments include the \textit{FairMLHealth} python package that focuses on algorithms and metrics for fairness in healthcare machine learning \cite{fairMLHealth}.

Additionally organizations like
Google and Amazon have set up AI ethics board emphasizing the importance of fairness in AI.
Given the extensive nature of the literature it is not possible to list all the relevant papers on fairness in machine learning here. We give an brief overview of papers that are most
relevant to the current manuscript. Researchers have proposed a
number of theories and methods to detect and measure fairness based on different definition of fairness in Machine
Learning. Zliobaite lists \cite{zliobaite2015survey} several statistical
methods and comparison functions to detect or measure different notions of fairness in Machine learning based on the
prediction or classification results, like Normalized Difference \cite{zliobaite2015survey} which is normalized mean difference
for binary classification used to quantify the difference between groups of people. 

Martinez et. al. explore fairness in healthcare in the context of risk-disparity among subpopulations. \cite{martinez2019fairness}.
Tramer et al. \cite{tramer2017fairtest}
has introduced unwarranted associations (UA) framework to
detect the fairness issue in data-driven applications by investigating associations between application outcomes and sensitive user attributes. There are also some detection methods
developed for specific problems or algorithms, like the detection methods for ranking algorithm. Corbett-Davies give a comprehensive survey of fairness in machine learning \cite{corbett2018measure}. Lastly, Ahmad et al. survey the field of healthcare AI within the context of fairness and describe the limitations and challenges of fairness in the healthcare domain \cite{ahmad2020fairness}.

\begin{table}[]
\begin{tabular}{l|l|l|l|l|l}
                           & \textbf{F}         & dataset & Male   & Female & var       \\ \hline
\multirow{2}{*}{AUC} & 1    & 0.6831         & 0.6908 & 0.6724 & 0.0001693 \\
                           & 0 & 0.6837         & 0.6887 & 0.6769 & 0.0000696 \\ \hline
\multirow{2}{*}{Precision} & 1    & 0.7156         & 0.7225 & 0.7046 & 0.0001602 \\
                           & 0 & 0.7274         & 0.7306 & 0.723  & 0.0000289 \\ \hline
\multirow{2}{*}{Recall}    & 1    & 0.755          & 0.7582 & 0.7506 & 0.0000289 \\
                           & 0 & 0.7563         & 0.7590 & 0.7526 & 0.0000205 \\ \hline
\multirow{2}{*}{F1}  & 1    & 0.6625         & 0.6682 & 0.6549 & 0.0000884 \\
                           & 0 & 0.6641         & 0.6683 & 0.6585 & 0.0000480
\end{tabular}
\caption{Performance of random forest models trained with
and without the protected feature \textbf{F} (gender) on LOS dataset}
  \label{tab:rf}
\end{table}

\section{Dimensions of Fairness in Machine Learning}
Even with the presence of competing notions of fairness
in machine learning, it is still possible to describe the various definitions of fairness in terms of the machine learning
pipeline. At a high level one can talk about three orthogonal
dimensions of data, algorithms and metrics. In this paper, we
focus our experiments on each of these dimensions. Each of
these can be described as follows:
\begin{itemize}
    \item \textit{Fairness in dataset:} Problems in fairness may be related
to the attributes of dataset, for instance, when one or several categories are under-represented, when the dataset is
outdated or incomplete \cite{ahmad2020fairness} \cite{crawford2013hidden}, or when the dataset inherits unintentional perpetuation and promotion
of historical biases, the machine learning methods may
learn these disparateness and unfairness and thus end
up operationalizing unfair outcomes. Unbiasing in this
case requires techniques that appropriately handle the data
while leaving the rest of the machine learning pipeline
intact. In this paper we address unfairness in dataset by
applying oversampling to the protected features and then
determining how does that change the results of the predictive models.
\item \textit{Fairness in model/algorithm:} Problems related to the design or even the choice of the machine learning model or
algorithm. Poorly designed matching systems, inappropriate choice of features, and assumptions like correlation
necessarily implies causation could all give rise to issues
related to unfairness. In certain contexts, one could even
state that just as data is not neutral, algorithms are also
not neutral. We explore this dimension in this paper by
considering how the performance and fairness of
different algorithms vary for different classification tasks
in healthcare.
    \item \textit{Fairness in metrics/results:} Problems related to the effect
of, and choice of metrics on measuring fairness. Additionally, problems related to unbiasing results from machine
learning models. The main approaches in this domain are
focused on post-processing solutions to rectify results after the fact. In this paper we focus on determining thresholds for models which can be then used to create a more
equitable outcome for protected features described below.
\end{itemize}

\section{Protected Features}
Protected or sensitive features are the features which can potentially be used to discriminate against populations e.g.,
race, gender, religion, ethnicity, caste, sexual orientation. It
may be the case that the use of such features may lead
to improvement in predictive performance in the machine
learning models but could also likely lead to discrimination.
Thus, the non-use of protected features in machine learning
models is recommended. Protected features can be divided
into two broad categories of features:

\begin{itemize}
    \item Known protected features: The attributes that are already protected by the law, e.g., the Equality Act of 2010 \cite{zliobaite2015survey} such as age, gender, disability, race, etc.
    \item Unknown protected features: These are potentially protected features which are non-obvious. Some data analysis or prior experience may be required to determine what
constitutes such features. Consider the example of inferring race given a person’s last name and zip code which
may not appear to be protected features at first. It has
been demonstrated that it is possible to build machine
learning models that can infer race based on these characteristics \cite{zliobaite2015survey} are also sometimes referred to as proxy features.
\end{itemize}
It is not always straightforward
to determine if a feature is sufficiently correlated with protected features and if we should include it in training As mentioned in the introduction, there are multiple notions of fairness in machine learning. In this paper is classification parity, which is the
performance parity of machine learning models with respect
to the protected features. Classification parity for protected
features can be defined as follows:
Protected Feature Classification Parity: Given a dataset
$D$ with $V$ attributes and a subset of protected features $\rho \subset
V$, for any given categorical attribute $v_i \in V$ with $C$ classes,
if the performance of a predictive model for Algorithm $A_q$ is
$\Phi (A_q, D, v_{ij} )$ for class $c_j \in C$ then the following condition
should hold for performance parity.
\begin{equation}
    \Phi(A_q, D, v_{ij} ) - \Phi(A_q, D, v_{ik}) < \xi, \forall~j \neq k
\end{equation}
where $\xi$ is a threshold corresponding to a relatively small
difference in performance of the various classes of the performance metrics. The performance metric can be any classification metric, for example precision, recall, AUC, F-Score, or Brier
Score. In other words, each class of the protected features should have similar performance on the quantitative
assessment metrics of relevance. For instance, consider the
case of models that are used to assess a criminal defendant’s
probability of becoming a recidivist i.e., a re-offending criminal. Such algorithms have been increasingly used across
the nation by probation officers, judges and parole officers
but studies have shown that these algorithms in general have
very different predictive performance across different racial
and ethnic groups.

\section{Experiments and Results}
\subsection{Dataset}
We employ data from three publicly available healthcare related datasets to study the effect of thresholds on fairness.
These are as follows:
\begin{itemize}
    \item MIMIC An ICU related dataset which has been extensively used in literature to study a number of prediction
problems in healthcare. The data considered consisted of
46,630 rows and 212 feature. In this paper we focus on
the problem of predicting length of stay at the time of admission to a medical facility. \cite{johnson2016mimic}
    \item Thyroid Disease Dataset Thyroid disease records supplied by the Garavan Institute in Australia. The data consisted of 3,772 rows and 29 features. The problem of predicting the presence or absence of the Thyroid disease is
addressed. \cite{dheeru2017karra}.
    \item Pima Indians Diabetes Dataset (PIMA) Dataset from
the National Institute of Diabetes and Digestive and Kidney Diseases which mainly consists of diagnostic measurements \cite{smith1988using}. The datset consisted of 768
rows and 9 features. We focus on the problem of predicting the presence or absence of diabetes in patients.
\end{itemize}
The target variables in all these cases are nominal, and thus
we pose these problems as classification problems. We note
that while the experiments outlined here were performed on
all three datasets, because of limitations in space we only
report results of a subset experiments because of limitations
of space.

\begin{figure} [t]
 \includegraphics[height=5cm, width=7.7cm]{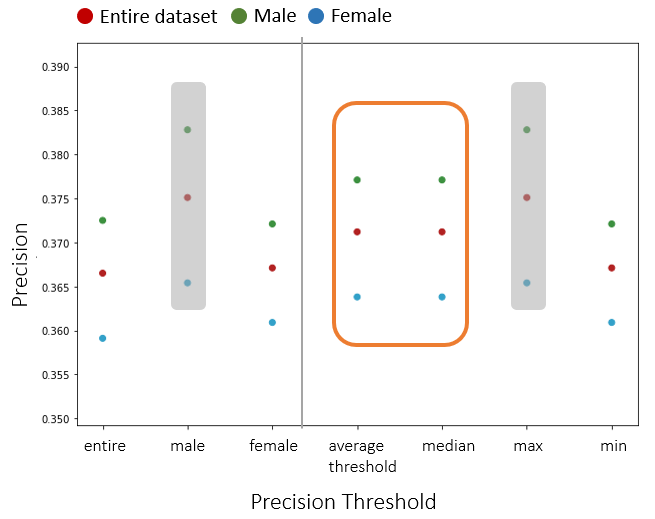}
  \caption{Precision thresholds for the overall dataset and the
constituent classes: male, female. The boundaries of fairness
thresholds are marked by orange rectangles}
  \label{fig:threshhold}
\end{figure}
\begin{table*}[]
\begin{tabular}{l|l|l|ll|ll|ll|ll|ll}
\multirow{2}{*}{Metric} & \multirow{2}{*}{$\sigma$} & \multirow{2}{*}{F} & \multicolumn{2}{c|}{\multirow{2}{*}{Entire dataset}} & \multicolumn{6}{c|}{Age}                                                                   & \multicolumn{2}{c}{\multirow{2}{*}{Variance}} \\ \cline{6-11}
                        &                           &                    & \multicolumn{2}{c|}{}            & \multicolumn{2}{c|}{xx-40} & \multicolumn{2}{c|}{41-70}       & \multicolumn{2}{c|}{71-xx} & \multicolumn{2}{c}{}        \\ \hline
                        & 0                         & 1                  &                                  & 0.9988            &              & 0.9961      &                         & 0.9997 &              & 0.9990      &                              & 3.60E-06        \\
AUC                     &                           & 0                  & -0.0023                          & 0.9965            & -0.0065      & 0.9896      & -0.0009                 & 0.9988 & -0.0010      & 0.998       & 6.2778                       & 2.62E-05        \\
                        & 1                         & 1                  &                                  & 0.9986            &              & 0.9955      &                         & 0.9996 &              & 0.9989      &                              & 4.90E-06        \\
                        &                           & 0                  & -0.0020                          & 0.9966            & -0.0055      & 0.99        & -0.0007                 & 0.9989 & -0.0011      & 0.9978      & 3.7551                       & 2.33E-05        \\ \hline
                        & 0                         & 1                  & -                                & 0.9747            &              & 0.9542      &                         & 0.9840 &              & 0.9771      &                              & 2.42E-04        \\
Precision               &                           & 0                  & -0.0079                          & 0.967             & -0.0258      & 0.9296      & \textcolor{red}{0.0027} & 0.9867 & -0.0035      & 0.9737      & 2.6932                       & 0.0008956       \\
                        & 1                         & 1                  &                                  & 0.9745            &              & 0.9553      &                         & 0.9856 &              & 0.9715      &                              & 2.31E-04        \\
                        &                           & 0                  & -0.0136                          & 0.9612            & -0.0279      & 0.9286      & -0.0047                 & 0.981  & -0.0132      & 0.9587      & 2.0013                       & 0.0006918       \\ \hline
                        & 0                         & 1                  &                                  & 0.9768            &              & 0.9587      &                         & 0.9858 &              & 0.9826      &                              & 2.19E-04        \\
Recall                  &                           & 0                  & -0.0074                          & 0.9696            & -0.0227      & 0.9369      & \textcolor{red}{0.0014} & 0.9872 & -0.0058      & 0.9769      & 2.2266                       & 0.0007076       \\
                        & 1                         & 1                  &                                  & 0.9768            &              & 0.9599      &                         & 0.9872 &              & 0.9769      &                              & 1.91E-04        \\
                        &                           & 0                  & -0.0117                          & 0.9654            & -0.0252      & 0.9357      & -0.0050                 & 0.9823 & -0.0080      & 0.9691      & 2.0220                       & 0.000576        \\ \hline
                        & 0                         & 1                  &                                  & 0.9750            &              & 0.9560      &                         & 0.9842 &              & 0.9789      &                              & 2.25E-04        \\
F1                      &                           & 0                  & -0.0080                          & 0.9672            & -0.0254      & 0.9317      & \textcolor{red}{0.0019} & 0.9861 & -0.0056      & 0.9734      & 2.6069                       & 0.0008112       \\
                        & 1                         & 1                  &                                  & 0.9748            &              & 0.9572      &                         & 0.9854 &              & 0.9713      &                              & 1.98E-04        \\
                        &                           & 0                  & -0.0131                          & 0.962             & -0.0277      & 0.9307      & -0.0051                 & 0.9804 & -0.0093      & 0.9623      & 2.1817                       & 0.0006303      
\end{tabular}
\caption{Performance of methods trained with and without the important feature (Age) related (proxy) to the protected features on Thyroid dataset, where the increase in performance are marked as red. Here $\sigma$ corresponds to whether sampling was done or not}
  \label{tab:ThyroidAge}
\end{table*}
\section{Model Assessment}
A standard machine learning pipeline consists of data collection, data pre-processing, feature selecting, algorithm selection, model training, model selection and model evaluation. In this paper, we focus on the data pre-processing, algorithm selection and model evaluation part of fairness in
machine learning. We propose a general way to assess classification parity for fairness in machine learning and determine the optimal thresholds for creating fair and unbiased
machine learning models. Specifically, for each protected
feature, we measured fairness by comparing the predictive
performance of each category for the feature. Afterwards we
compared the optimal threshold for predictive performance
chosen based on the performance of each category. Next,
we determined a fairness threshold which corresponded to
optimal outcome across classes of the protected features.
The fairness threshold allows us to create classification models with fair outcomes without significant drop in predictive
performance. We measure predictive performance in terms
of standard classification metrics like AUC, precision, recall
and F1 score.

The optimal threshold was found based on Youden’s index \cite{youden1950index}:
\begin{equation}
    J = sensitivity + specificity - 1 
\end{equation}
Youden’s index has been used as the measure of diagnostic
effectiveness. It could also be used for selecting the optimal
cut-point on ROC-curve \cite{schisterman2005optimal}, which
corresponds to the optimal threshold we desire. Additionally we measured performance of the predictive models at
the level of data and features i.e., training without the protected features, training without the important features related to the protected feature to help reduce the unfairness,
and sampling to balance the size of each category on training dataset. Due to space limitations we limit the analysis
of algorithmic performance to the following three popular,
well-studied and well-applied algorithms in the healthcare
domain: Logistic Regression, Random Forest and XGBoost
(eXtreme Gradient Boosting).

\begin{table*}[]
\begin{tabular}{l|l|l|ll|ll|ll|ll}
\multirow{2}{*}{Metric} &  \multirow{2}{*}{$\sigma$}  & \multirow{2}{*}{F} & \multicolumn{2}{c|}{\multirow{2}{*}{Entire dataset}} & \multicolumn{4}{c|}{Gender}                                                                             & \multicolumn{2}{c}{\multirow{2}{*}{Variance}} \\ \cline{6-9}
                                 &   &                             & \multicolumn{2}{c|}{}                                         & \multicolumn{2}{c}{male} &  \multicolumn{2}{c|}{female} & \multicolumn{2}{c}{}                                   \\ \hline
          & 0 & 1        &         & 0.9988  &         & 0.9994   &         & 0.9984 &          & 4.00E-07  \\
AUC       &   & 0 & -0.0028 & 0.996   & -0.0020 & 0.9974   & -0.0035 & 0.9949 & 6.2500   & 2.90E-06  \\
          & 1 & 1        &         & 0.9977  &         & 0.9972   &         & 0.9978 &          & 2.00E-07  \\
          &   & 0 & -0.0025 & 0.9952  & \textcolor{red}{0.0002}  & 0.9974   & -0.0040 & 0.9938 & 32.5000  & 6.70E-06  \\
          & 0 & 1        &         & 0.9713  &         & 0.9816   &         & 0.9644 &          & 1.47E-04  \\ \hline
Precision &   & 0 & -0.0096 & 0.962   & -0.0001 & 0.9815   & -0.0122 & 0.9526 & 1.8366   & 0.0004167 \\
          & 1 & 1        &         & 0.9665  &         & 0.9795   &         & 0.9577 &          & 2.39E-04  \\
          &   & 0 & -0.0138 & 0.9532  & -0.0013 & 0.9782   & -0.0187 & 0.9398 & 2.0951   & 0.0007388 \\
          & 0 & 1        &         & 0.9743  &         & 0.9876   &         & 0.9672 &          & 2.09E-04  \\ \hline
Recall    &   & 0 & -0.0070 & 0.9675  & -0.0004 & 0.9872   & -0.0087 & 0.9588 & 0.9410   & 0.0004047 \\
          & 1 & 1        &         & 0.9711  &         & 0.9876   &         & 0.9623 &          & 3.21E-04  \\
          &   & 0 & -0.0103 & 0.9611  & \textcolor{red}{0.0008}  & 0.9884   & -0.0138 & 0.949  & 1.4183   & 0.0007758 \\
          & 0 & 1        &         & 0.9719  &         & 0.9838   &         & 0.9648 &          & 1.81E-04  \\ \hline
F1        &   & 0 & -0.0094 & 0.9628  & -0.0004 & 0.9834   & -0.0118 & 0.9534 & 1.4814   & 0.0004479 \\
          & 1 & 1        &         & 0.9675  &         & 0.9828 - &         & 0.9586 &          & 2.94E-04  \\
          &   & 0 & -0.0124 & 0.9555  & \textcolor{red}{0.0004}  & 0.9832   & -0.0168 & 0.9425 & 1.8206   & 0.0008287
\end{tabular}
\caption{Performance of methods trained with and without the important feature (Gender) related (proxy) to the protected features on 
Thyroid dataset, where the increase in performance are marked as red. Here $\sigma$ corresponds to whether sampling was done or not}
  \label{tab:Thyroidgender}
\end{table*}

\section{Classification Parity and Fairness}
\textbf{Definition:} Given a classification task $T$, dataset $D$ with $m$
classes $C_1, C_2, C_3, ..., C_m$, evaluation metric $\Theta$, the fairness
threshold is defined as follows:
\begin{equation}
    argmin(\sigma^2(\Theta(Ci), \Theta(Cj ), \Theta(D))), \forall C_j , C_j \in C
\end{equation}
In other words, for a given classification task and protected feature with m classes the fairness threshold is the
threshold where the variance in the predictive performance
of an algorithm in minimized with respect to each of the
m classes as well as the overall dataset. To illustrate this
concept, consider Figure \ref{fig:threshhold} which shows the performance of
a Random Forest model for the Length of Stay prediction
task. Here the protected variable is gender and performance
is measured in terms of precision. The performance is given
for the two classes of gender (i.e., male and female) for this
dataset as well as the overall performance. The x-axis shows
the class or formula with respect to which precision is maximized, and the y-axis shows the precision of the model.

Consider the threshold that are used to maximize performance for female, the threshold would also correspond to some non-optimal performance for male and also for the overall population. Similarly, consider the threshold for fairness when the performance of
the predictive model is most fair for male. The two other
values in the graph are values that are obtained if the fairness threshold for the male population is used for the female
population as well as the overall population. Now consider
the thresholds given on the right in the figure, these are the
thresholds that are computed as the aggregate (average, min,
max etc) of the thresholds for the protected classes as well as
the overall population. We computed the performance when
the thresholds are chosen based on using minimum, maximum, average and median threshold are used. From Figure \ref{fig:threshhold} it is clear that the best results are obtained when the average or the minimum threshold is used.
Table \ref{tab:rf} shows the results of prediction for the length
of stay prediction problem along with the variance of the
performance of all the categories, which are ’male’ and ’female’ in this example. The variance in this case quantifies the difference of performance. The lesser the variance is, the
lesser is the difference in predictive performance. This also
implies that the models are more fair.
To find the threshold for fairness, we set the optimal
threshold for entire dataset, optimal thresholds for each category, and the average, median, maximum and minimum
value of the thresholds of all the categories as the candidates.
And then, we computed the difference of performance corresponding to each threshold, including precision, recall and
F1 score, across each category, and chose the one with minimum difference as the fairness threshold. To help analyze
effect on performance of choosing different threshold, we
plotted performance boundary, including precision boundary, recall boundary and F1 score boundary, for each fairness
threshold candidate.
The performance boundary shows that best performance
does not mean fair. For instance, the threshold based on male
which is also the maximum threshold in Figure \ref{fig:threshhold} has the
highest precision, but the it also has maximum difference,
which means the performance corresponding to this threshold is the least fair one. Furthermore, model with fair outcomes can have comparable performance. For instance, the
fairness threshold in Table \ref{tab:rf}  performs even a bit better than
the performance corresponding to optimal threshold for entire dataset, which we normally care about.

\section{Effect of Removal of Protected Features}
To make the performance of the protected feature’s each category similar to each other, removing the protected features
so that they could not influence the performance directly is
one apparent choice. Tables \ref{tab:LOSage},\ref{tab:LOSgender} and \ref{tab:LOSrace} in the appendix gives the performance of methods
training with and without protected features on LOS dataset for age, gender and race respectively. The increase in performance are marked as red and the
decrease in variance of the performance of all the categories
are marked as blue.

One thing to note is that the model trained without the
protected feature, gender, has more fair performance than
the one trained with all the features. This also implies that
underlying model was most likely using gender in its prediction. One can also find the optimal threshold for entire
dataset and the ones for all the categories. We note that in
this particular example the difference in the variance of the
models is not statistically significant. We found similar results for the other two models. We however emphasize that
the current models are for demonstrating the feasibility of
the proposed methods and we plan to explore this further as
described in the future work section below.

With the above disclaimer, one can still observe that there
are certain differences in the performance of the algorithms
which can be used to design experiments and analysis in the
future. In summary, the results show that:
\begin{itemize}
    \item The variance of the performance is more likely to decrease 10$\%$ - 30$\%$, sometimes over 60$\%$, which means
removal of protected features could help the performance
become more fair;
    \item Although it may appear that the performance is more
likely to improve with the inclusion of the protected feature, it mostly increases under 5$\%$, which could be interpreted as the performance shows insignificant
change.
\end{itemize}

\begin{table*}[]
\begin{tabular}{l|lll|lll}
                                & \multicolumn{3}{c|}{AUC}   & \multicolumn{3}{c}{AUC Variance} \\ \hline
                                & XGBoost & RF     & LR     & xgboost   & RF        & LR       \\ \hline
with 'race', before sampling    & 0.7066  & 0.6951 & 0.6557 & 0.000279  & 0.000215  & 0.000169 \\ \hline
without 'race', before sampling & 0.7069  & 0.6949 & 0.6552 & 0.000286  & 0.000214  & 0.000170 \\ \hline
with 'race', after sampling     & 0.6944  & 0.6887 & 0.6436 & 0.000441  & 0.000318  & 0.000331 \\ \hline
without 'race', after sampling  & 0.6918  & 0.6869 & 0.6431 & 0.000391  & 0.000305  & 0.000299
\end{tabular}
\caption{AUC scores and variance of AUC scores of all the categories of the experiments choosing race as the protected feature (RF = Random Forest, LR = Logistic Regression) for Diabetes Dataset}
  \label{tab:diabetes}
\end{table*}

\section{Effect of Removal of Proxy Features}
A further idea that we explore is the effect of the removal
of proxy features, especially ones with high importance
scores with respect to the predictive performance. This is to
ensure that the indirect influence of the protected features on the outcomes is not factored into the model and model
in general is fair. We define important features as follows,
given n features the important features are top k features
rank sorted by their important scores. The importance of a
feature is calculated by how much the performance measure
would improve on each attribute split and weighted by the
number of observations the node is responsible for while
training with Gradient Boosting method \cite{dash1997feature} \cite{xu2014gradient}.
Since we more interested in the proxy features, we only consider the important features that are highly correlated with
the protected features.
Table \ref{tab:Thyroidgender} and Table \ref{tab:ThyroidAge} gives the performance of models trained on the Thyroid dataset, with and
without the important features related to the protected features for gender and age respectively. For the important feature, we
consider k = 5. The main takeaways can be
summarized as follows: And it shows that:
\begin{itemize}
    \item The variance of the performance does not decrease if the
important features are removed.
    \item As expected, the performance of the models decreases when the important features are removed.
\end{itemize}
These observations imply that removing related important features is not helpful for either improvement in predictive performance or for fairness in general.

\section{Effect of Sampling}
The distribution of classes in most protected features are imbalanced. Consequently, most problems related to class imbalance in supervised learning are also prominent problems in this
domain. One way to mitigate the problem of class imbalance in supervised learning is to over-sample the under-represented classes. We considered the distribution
of the protected feature ’age’ and another protected feature ’race’ in the LOS dataset as examples. We observe that minority
populations like African Americans and Asians are underrepresented in the race feature. Similarly, pediatric patients
are under-represented in the age feature. To reduce the influence of under representation of the several categories, we
oversampled training dataset to balance the size of each
category. We observe that the result of determining the optimal threshold before and after sampling is that the value of
the optimal threshold changes once sampling is done.

\section{Fairness in Methods}
By comparing the AUC scores and the variance of AUC of the three methods, we could see the rank of the
three methods based on AUC score is:\\ \\
Logistic regression $<$ Random Forest $<$ XGBoost\\ \\
And the rank based on the variance in AUC is:\\
Logistic regression $<$ Random Forest $<$ XGBoost(Race)\\ \\
This could lead us to the conclusion that an accurate model
does not necessarily imply fair outcomes. The summary results for the experiments for the Diabetes datasets are given
in Table \ref{tab:diabetes}.

\section{Conclusion}
Identifying unfairness is a challenging task. In this work, we
first compared the predictive performance across protected features, and use the variance of the performance as a criteria to measure fairness. Second, we determined the optimal
thresholds chosen based on each category. We also explored
several ways to address unfairness. Additionally, we
trained models without the protected features, which could
help reduce unfairness but did not cause a 
drop in performance. The second method focused on the data
dimension, which is critical in a machine learning process 
because characteristics like under-representation could
be inherited or even exacerbated in machine learning process. When the size of dataset is small, or the one or several
category is under-represented, one could sample the training dataset first to balance the size of each category. Finally,
when one has detected unfairness and preferred to address it
without training the model again, one could find fairness
threshold, which could make the performance more fair but
also comparable.

Furthermore, comparison among AUC scores of different models lead us to the conclusion that an accurate
model does not necessarily imply fairness. Models with
higher accuracy may have less fair outcomes. We note that
the fairness measurement obtained for best results for each prediction
problem do not necessarily correspond to the best possible
theoretical results. This is work in progress, and in follow up
to this work we plan to explore theoretical aspects of fairness threshold. The methods and experiments described in
this paper are part of a proof of concept to test the efficacy of
methods to detect unfairness in machine learning in healthcare use cases. Our plan is to incorporate insights gleaned
from this exploratory analysis into a production-deployed
 system at a large scale healthcare system in the United
States.

\bibliographystyle{aaai}
\bibliography{biblio}

\appendix
\begin{table*}[]
\begin{tabular}{l|l|ll|ll|ll|ll}
\multicolumn{1}{c|}{} & \multicolumn{1}{c|}{\multirow{3}{*}{Features}} & \multicolumn{2}{c|}{\multirow{2}{*}{entire dataset}}  & \multicolumn{4}{c|}{gender}                                                                                  & \multicolumn{2}{c}{\multirow{2}{*}{Variance}} \\ \cline{5-8}
\multicolumn{1}{c|}{} & \multicolumn{1}{c|}{}                          & \multicolumn{2}{c|}{}        & \multicolumn{2}{c|}{male}                             & \multicolumn{2}{c|}{female}                          & \multicolumn{2}{c}{}            \\ \cline{3-10} 
\multicolumn{1}{c|}{} & \multicolumn{1}{c|}{}                          & \multicolumn{1}{c|}{no gender} & \multicolumn{1}{c|}{-} & \multicolumn{1}{c|}{no gender} & \multicolumn{1}{c|}{-} & \multicolumn{1}{c|}{no gender} & \multicolumn{1}{c|}{-} & \multicolumn{1}{c|}{no gender}     & -           \\ \hline
\multicolumn{10}{c}{\textbf{XGBoost}}   \\ \hline
AUC                   & with 'gender                                   & -                            & 0.7075                 & -                            & 0.7139                & -                            & 0.0993                & -                                & 0.000107    \\
                      & without 'gender                                & -0.0013                      & 0.7000                 & -0.0011                      & 0.7131                & -0.0014                      & 0.0983                & 0.0206                           & 0.000109    \\
                      & remove all                                     & -0.0017                      & 0.7063                 & -0.0018                      & 0.7126                & -0.0020                      & 0.0979                & 0.0124                           & 0.000108    \\
Precision             & with 'gender                                   & -                            & 0.7280                 & -                            & 0.7317                & -                            & 0.7252                & -                                & 0.000021    \\
                      & without 'gender                                & -0.0008                      & 0.728                  & -0.0011                      & 0.7309                & -0.0001                      & 0.7251                & \textcolor{blue}{-0.1952}                          & 0.000017    \\
                      & remove all                                     & -0.0004                      & 0.7283                 & -0.0023                      & 0.73                  & \textcolor{red}{0.0019}                       & 0.7200                & \textcolor{blue}{-0.7186}                          & 0.000006    \\
Recall                & with 'gender                                   & -                            & 0.7029                 & -                            & 0.7055                & -                            & 0.7595                & -                                & 0.000018    \\
                      & without 'gender                                & -0.0003                      & 0.7027                 & -0.0003                      & 0.7053                & -0.0003                      & 0.7993                & \textcolor{blue}{-0.0131}                          & 0.000018    \\
                      & remove all                                     & -0.0001                      & 0.7028                 & -0.0008                      & 0.7649                & \textcolor{red}{0.0007}                       & 0.76                  & \textcolor{blue}{-0.3437}                          & 0.000012    \\
F1                    & with 'gender                                   & -                            & 0.701                  & -                            & 0.7030                & -                            & 0.0970                & -                                & 0.000018    \\
                      & without 'gender                                & -0.0009                      & 0.7004                 & -0.0004                      & 0.7033                & -0.0016                      & 0.0985                & 0.3077                           & 0.000023    \\
                      & remove all                                     & -0.0006                      & 0.7006                 & -0.0016                      & 0.7025                & \textcolor{red}{0.0003}                       & 0.0978                & \textcolor{blue}{-0.3841}                          & 0.000011    \\ \hline
\multicolumn{10}{c}{\textbf{Random Forest}}   \\ \hline

AUC                   & with 'gender                                   & -                            & 0.094                  & -                            & 0.0975                & -                            & 0.6895                & -                                & 0.000032    \\
                      & without 'gender                                & -0.0003                      & 0.0938                 & \textcolor{red}{0.0017}                       & 0.0987                & -0.0032                      & 0.6873                & 1.0429                           & 0.000005    \\
                      & remove all                                     & -0.0035                      & 0.0916                 & -0.0017                      & 0.0963                & -0.0061                      & 0.6853                & 0.9195                           & 0.000061    \\
Precision             & with 'gender                                   & -                            & 0.7305                 & -                            & 0.7358                & -                            & 0.7254                & -                                & 0.000055    \\
                      & without 'gender                                & -0.0015                      & 0.7294                 & -0.0001                      & 0.7357                & -0.0052                      & 0.7216                & 0.3187                           & 0.000099    \\
                      & remove all                                     & -0.0042                      & 0.7274                 & -0.0049                      & 0.7322                & -0.0043                      & 0.7223                & \textcolor{blue}{-0.1171}                          & 0.000048    \\
Recall                & with 'gender                                   & -                            & 0.7008                 & -                            & 0.7639                & -                            & 0.7505                & -                                & 0.000028    \\
                      & without 'gender                                & -0.0001                      & 0.7007                 & \textcolor{red}{0.0003}                       & 0.7641                & -0.0005                      & 0.7561                & 0.1768                           & 0.000032    \\
                      & remove all                                     & -0.0004                      & 0.7005                 & -0.0004                      & 0.7630                & -0.0001                      & 0.7564                & \textcolor{blue}{-0.0373}                          & 0.000027    \\
F1                    & with 'gender                                   & -                            & 0.0842                 & -                            & 0.0889                & -                            & 0.6778                & -                                & 0.000062    \\
                      & without 'gender                                & \textcolor{red}{0.0009}                       & 0.0848                 & \textcolor{red}{0.0009}                       & 0.6895                & \textcolor{red}{0.0010}                       & 0.6785                & \textcolor{blue}{-0.0216}                          & 0.000000    \\
                      & remove all                                     & \textcolor{red}{0.0037}                       & 0.6867                 & \textcolor{red}{0.0032}                       & 0.0911                & \textcolor{red}{0.0044}                       & 0.6808                & \textcolor{blue}{-0.1391}                          & 0.000053    \\ \hline
\multicolumn{10}{c}{\textbf{Logistic Regression}}   \\ \hline
AUC                   & with 'gender                                   & -                            & 0.0538                 & -                            & 0.0615                & -                            & 0.6435                & -                                & 0.000163    \\
                      & without 'gender                                & -0.0003                      & 0.0536                 & -0.0002                      & 0.0614                & 0.0002                       & 0.6436                & \textcolor{blue}{-0.0340}                          & 0.000158    \\
                      & remove all                                     & -0.0012                      & 0.053                  & \textcolor{red}{0.0003}                       & 0.0617                & -0.0022                      & 0.6421                & 0.1089                           & 0.000191    \\
Precision             & with 'gender                                   & -                            & 0.705                  & -                            & 0.711                 & -                            & 0.097                 & -                                & 0.000098    \\
                      & without 'gender                                & \textcolor{red}{0.0003}                       & 0.7052                 & -0.0011                      & 0.7102                & \textcolor{red}{0.0023}                       & 0.0986                & \textcolor{blue}{-0.3165}                          & 0.000067    \\
                      & remove all                                     & \textcolor{red}{0.0010}                       & 0.7057                 & \textcolor{red}{0.0001}                       & 0.7111                & \textcolor{red}{0.0024}                       & 0.0987                & \textcolor{blue}{-0.2007}                          & 0.000078    \\
Recall                & with 'gender                                   & -                            & 0.633                  & -                            & 0.6426                & -                            & 0.0202                & -                                & 0.000250    \\
                      & without 'gender                                & \textcolor{red}{0.0009}                       & 0.6336                 & \textcolor{red}{0.0100}                       & 0.649                 & -0.0114                      & 0.6131                & 1.5726                           & 0.000644    \\
                      & remove all                                     & \textcolor{red}{0.0033}                       & 0.6351                 & \textcolor{red}{0.0154}                       & 0.0525                & -0.0137                      & 0.6117                & 23232                            & 0.000832    \\
F1                    & with 'gender                                   & -                            & 0.0555                 & -                            & 0.0644                & -                            & 0.6435                & -                                & 0.000218    \\
                      & without 'gender                                & \textcolor{red}{0.0008}                       & 0.056                  & \textcolor{red}{0.0074}                       & 0.0093                & -0.0092                      & 0.6375                & 1.3020                           & 0.000503    \\
                      & remove all                                     & \textcolor{red}{0.0027}                       & 0.0573                 & \textcolor{red}{0.0119}                       & 0.6723                & -0.0110                      & 0.6364                & 1.9391                           & 0.000642   
\end{tabular}
\caption{Performance of methods trained with and without the important feature (Gender) related (proxy) to the protected features on LOS dataset, where the increase in performance are marked as red and decrease is marked by blue}
  \label{tab:LOSgender}
\end{table*}

\begin{table*}[]
\begin{tabular}{l|l|ll|ll|ll|ll|ll}
\multicolumn{1}{c|}{} & \multicolumn{1}{c|}{\multirow{2}{*}{Features}} & \multicolumn{2}{c|}{\multirow{2}{*}{entire dataset}} & \multicolumn{6}{c|}{age}                                                                                                                                               & \multicolumn{2}{c}{\multirow{2}{*}{Variance}} \\  \cline{5-10}
\multicolumn{1}{c|}{} & \multicolumn{1}{c|}{}                          & \multicolumn{2}{c|}{}                                & \multicolumn{2}{c|}{xx-40}                            & \multicolumn{2}{c|}{41-70}                             & \multicolumn{2}{c|}{71-xx}                            & \multicolumn{2}{c}{}                          \\  \cline{3-12}
\multicolumn{1}{c|}{} & \multicolumn{1}{c|}{}                          &  \multicolumn{1}{c|}{no 'age}           & \multicolumn{1}{c|}{-}           & \multicolumn{1}{c|}{no 'age} & \multicolumn{1}{c|}{-} & \multicolumn{1}{c|}{no 'age'} & \multicolumn{1}{c|}{-} & \multicolumn{1}{c|}{no 'age} &  \multicolumn{1}{c|}{.} & \multicolumn{1}{c|}{no 'age}    &             \\ \hline
\multicolumn{12}{c}{\textbf{XGBoost}}                                                                                                                                                                                                                                                                                                                          \\ \hline 
                      & 1                                      & -                 & 0.7045                           & -                            & 0.7424                 & -                             & 0.7135                 & -                            & 0.0800                 & -                               & 0.000779    \\
AUC                   & 0                                   & -0.0017           & 0.7033                           & \textcolor{red}{0.0000}                       & 0.7424                 & -0.0000                       & 0.7131                 & -0.0015                      & 0.0850                 & 0.0359                          & 0.000807    \\
                      & $\Delta$                                     & -0.0009           & 0.7039                           & \textcolor{red}{0.0016}                     & 0.7430                 & \textcolor{red}{0.0004}                        & 0.7138                 & -0.0010                      & 0.0859                 & 0.0093                          & 0.000833    \\ \hline 
                      & 1                                      & -                 & 0.7208                           & -                            & 0.7835                 & -                             & 0.7454                 & -                            & 0.7005                 & -                               & 0.001482    \\ 
Precision             & 0                                   & \textcolor{red}{0.0007}            & 0.7273                           & -0.0032                      & 0.781                  & \textcolor{red}{0.0030}                        & 0.7470                 & \textcolor{red}{0.0004}                       & 0.7008                 & \textcolor{blue}{-0.0082}                         & 0.001381    \\
                      & $\Delta$                                     & \textcolor{red}{0.0021}            & 0.7283                           & -0.0043                      & 0.7801                 & \textcolor{red}{0.0034}                  & 0.7479                 & \textcolor{red}{0.0030}                       & 0.7080                 & \textcolor{blue}{-0.1350}                         & 0.001282    \\ \hline 
                      & 1                                      & -                 & 0.7022                           & -                            & 0.8131                 & -                             & 0.7781                 & -                            & 0.7433                 & -                               & 0.001216    \\ 
Recall                 & 0                                   & \textcolor{red}{0.0001}            & 0.7023                           & -0.0010                      & 0.8123                 & \textcolor{red}{0.0005}                        & 0.7785                 & \textcolor{red}{0.0003}                   & 0.7435                 & \textcolor{blue}{-0.0255}                         & 0.0011a5    \\
                      & $\Delta$                                     & \textcolor{red}{0.0007}            & 0.7627                           & -0.0015                      & 0.8119                 & \textcolor{red}{0.0008}                       & 0.7787                 & \textcolor{red}{0.0012}                       & 0.7442                 & \textcolor{blue}{-0.0576}                         & 0.001148    \\ \hline 
                      & 1                                      & -                 & 0.7                              & -                            & 0.7012                 & -                             & 0.7183                 & -                            & 0.678                  & -                               & 0.001730    \\ 
F1             & 0                                   & -0.0000           & 0.0990                           & -0.0012                      & 0.7003                 & -0.0010                       & 0.7176                 & -0.0003                      & 0.6778                 & \textcolor{blue}{-0.0162}                         & 0.001702    \\
                      & $\Delta$                                     & \textcolor{red}{0.0001}            & 0.7001                           & -0.0030                      & 0.7589                 & -0.0010                       & 0.7176                 & \textcolor{red}{0.0013}                       & 0.6789                 & \textcolor{blue}{-0.0746}                         & 0.001601    \\  \hline 

\multicolumn{12}{c}{\textbf{Random Forest}}                                                                                                                                                                                                                                                                                                                     \\ \hline 
AUC                       & 1                                      & -                 & 0.0921                           & -                            & 0.7385                 & -                             & 0.7008                 & -                            & 0.6727                 & -                               & 0.001090    \\
                  & 0                                   & -0.0027           & 0.0902                           & -0.0007                      & 0.738                  & -0.0051                       & 0.0972                 & -0.0007                      & 0.6722                 & 0.0119                          & 0.001103    \\
                      & $\Delta$                                     & -0.0055           & 0.6883                           & -0.0073                      & 0.7331                 & -0.0123                       & 0.0922                 & -0.0003                      & 0.6725                 & \textcolor{blue}{-0.1248}                         & 0.000954    \\ \hline 
                      & 1                                      & -                 & 0.7274                           & -                            & 0.7820                 & -                             & 0.7481                 & -                            & 0.7057                 & -                               & 0.001482    \\ 
Precision             & 0                                   & -0.0077           & 0.7218                           & -0.0121                      & 0.7731                 & -0.0050                       & 0.7439                 & -0.0081                      & 0.7                    & -0.0857                         & 0.001355    \\
                      & $\Delta$                                     & -0.0100           & 0.7201                           & -0.0079                      & 0.7704                 & -0.0099                       & 0.7407                 & -0.0111                      & 0.0979                 & 0.0432                          & 0.001546    \\ \hline 
                      & 1                                      & -                 & 0.70                             & -                            & 0.8121                 & -                             & 0.770                  & -                            & 0.7408                 & -                               & 0.001271    \\ 
Recall                 & 0                                   & -0.0016           & 0.7588                           & -0.0037                      & 0.8091                 & -0.0000                       & 0.7755                 & -0.0015                      & 0.7397                 & \textcolor{blue}{-0.0527}                         & 0.001204    \\
                      & $\Delta$                                     & -0.0018           & 0.7580                           & -0.0020                      & 0.8105                 & -0.0012                       & 0.7751                 & -0.0022                      & 0.7392                 & 0.0000                          & 0.001271    \\ \hline 
                      & 1                                      & -                 & 0.0832                           & -                            & 0.750                  & -                             & 0.7032                 & -                            & 0.0579                 & -                               & 0.002413    \\ 
F1             & 0                                  & -0.0007           & 0.0827                           & -0.0040                      & 0.753                  & \textcolor{red}{0.0013}                        & 0.7041                 & -0.0011                      & 0.0572                 & \textcolor{blue}{-0.0497}                         & 0.002293    \\
                      & $\Delta$                                     & \textcolor{red}{0.0010}            & 0.0839                           & \textcolor{red}{0.0001}                       & 0.7561                 & \textcolor{red}{0.0031}                        & 0.7054                 & \textcolor{red}{0.0000}                       & 0.0579                 & 0.0000                          & 0.002413    \\ \hline 

\multicolumn{12}{c}{\textbf{Logistic Regression}}                                                                                                                                                                                                                                                                                                              \\ \hline 
                      & 1                                      & -                 & 0.0515                           & -                            & 0.083                  & -                             & 0.0007                 & -                            & 0.0340                 & -                               & 0.0005a0    \\ 
AUC                   & 0                                   & -0.0008           & 0.051                            & -0.0023                      & 0.0814                 & \textcolor{red}{0.0011}                        & 0.0614                 & -0.0003                      & 0.0344                 & \textcolor{blue}{-0.0495}                         & 0.000557    \\
                      & $\Delta$                                     & -0.0009           & 0.0509                           & -0.0107                      & 0.6757                 & \textcolor{red}{0.0014}                        & 0.0616                 & -0.0002                      & 0.6345                 & \textcolor{blue}{-0.2543}                         & 0.000437    \\ \hline
                      & 1                                      & -                 & 0.7047                           & -                            & 0.7638                 & -                             & 0.7199                 & -                            & 0.083                  & -                               & 0.001639    \\
Precision             & 0                                   & -0.0001           & 0.7046                           & -0.0041                      & 0.7007                 & \textcolor{red}{0.0022}                        & 0.7215                 & -0.0012                      & 0.0822                 & \textcolor{blue}{-0.0904}                         & 0.001540    \\
                      & $\Delta$                                     & \textcolor{red}{0.0007}            & 0.7052                           & -0.0022                      & 0.7021                 & \textcolor{red}{0.0019}                        & 0.7213                 & \textcolor{red}{0.0004}                       & 0.0833                 & \textcolor{blue}{-0.0531}                         & 0.001552    \\ \hline
                      & 1                                      & -                 & 0.031                            & -                            & 0.7027                 & -                             & 0.0717                 & -                            & 0.5825                 & -                               & 0.008113    \\  
Recall                 & 0                                   & \textcolor{red}{0.0021}            & 0.6323                           & -0.0229                      & 0.7452                 & \textcolor{red}{0.0138}                        & 0.081                  & \textcolor{red}{0.0010}                       & 0.5831                 & \textcolor{blue}{-0.1782}                         & 0.000007    \\
                      & $\Delta$                                     & \textcolor{red}{0.0040}            & 0.6335                           & -0.0220                      & 0.7459                 & \textcolor{red}{0.0165}                        & 0.0828                 & \textcolor{red}{0.0031}                       & 0.5843                 & \textcolor{blue}{-0.1818}                         & 0.000638    \\ \hline 
                      & 1                                      & -                 & 0.0539                           & -                            & 0.7027                 & -                             & 0.0894                 & -                            & 0.0009                 & -                               & 0.000081    \\ 
F1             & 0                                   & \textcolor{red}{0.0014}            & 0.0548                           & -0.0143                      & 0.7518                 & \textcolor{red}{0.0103}                        & 0.0905                 & \textcolor{red}{0.0008}                       & 0.0074                 & \textcolor{blue}{-0.1273}                         & 0.005307    \\
                      & $\Delta$                                     & \textcolor{red}{0.0032}            & 0.050                            & -0.0131                      & 0.7527                 & \textcolor{red}{0.0119}                        & 0.0970                 & \textcolor{red}{0.0026}                      & 0.0085                 & \textcolor{blue}{-0.1289}                          & 0.005297   
\end{tabular}
\caption{Performance of methods trained with and without the important feature (Age) related (proxy) to the protected features on 
LOS dataset, where the increase in performance are marked as red and decrease is marked by blue. Features = 1 implies inclusion of the feature; Features = 0 implies inclusion of the feature; Features = $\Delta$ implies 'remove all'}
  \label{tab:LOSage}
\end{table*}

\begin{landscape}
\centering
\begin{table}[]
\begin{tabular}{l|l|ll|ll|ll|ll|ll|ll|ll}
\multicolumn{1}{c|}{\multirow{3}{*}{}} & \multicolumn{1}{c|}{\multirow{3}{*}{F}} & \multicolumn{2}{c|}{\multirow{2}{*}{entire dataset}}  & \multicolumn{10}{c|}{Race}                                                                                                                                                                                                                                        & \multicolumn{2}{c}{\multirow{2}{*}{Variance}} \\ \cline{5-14}
\multicolumn{1}{c|}{}                  & \multicolumn{1}{c|}{}                          & \multicolumn{2}{c|}{}        & \multicolumn{2}{c}{White}                             & \multicolumn{2}{c|}{Black}                            & \multicolumn{2}{c|}{Hispanic/Latino}                   & \multicolumn{2}{c|}{Asian}       & \multicolumn{2}{c|}{Other}                            & \multicolumn{2}{c}{}            \\ \cline{3-16} 
\multicolumn{1}{c|}{}                  & \multicolumn{1}{c|}{}                          & \multicolumn{1}{c|}{no race} & \multicolumn{1}{c|}{-} & \multicolumn{1}{c|}{no race} & \multicolumn{1}{c|}{-} & \multicolumn{1}{c|}{no race} & \multicolumn{1}{c|}{-} & \multicolumn{1}{c|}{no race} & \multicolumn{1}{c|}{-} & \multicolumn{1}{c|}{no race} & \multicolumn{1}{c|}{-} & \multicolumn{1}{c|}{no race} & \multicolumn{1}{c|}{-} & \multicolumn{1}{c|}{no race}    & \multicolumn{1}{c}{-}           \\ \hline
\multicolumn{16}{c}{\textbf{XGBoost}} \\ \hline
AUC                                    & 1                                     & -                            & 0.7000                 &                              & 0.7005                 & -                            & 0.0935                 & -                            & 0.7207                 &         & 0.7042                 & -                            & 0.7340                 & -                                & 0.000279    \\
                                       & 0                                        & \textcolor{red}{0.0004}                       & 0.7009                 & \textcolor{red}{0.0006}                       & 0.7009                 & -0.0009                      & 0.0929                 & -0.0010                      & 0.72                   & -0.0003 & 0.704                  & \textcolor{red}{0.0007}                       & 0.7351                 & \textcolor{blue}{0.0251}                           & 0.000296    \\
                                       & $\Delta$                                     & -0.0024                      & 0.7049                 & -0.0019                      & 0.0992                 & -0.0022                      & 0.092                  & -0.0105                      & 0.7131                 & \textcolor{red}{0.0044}  & 0.7073                 & -0.0042                      & 0.7315                 & -0.1828                          & 0.000228    \\ \hline
Precision                              & 1                                      & -                            & 0.7298                 & -                            & 0.7233                 & -                            & 0.7479                 & -                            & 0.7730                 & -       & 0.741                  & -                            & 0.7311                 & -                                & 0.000374    \\
                                       & 0                                   & \textcolor{red}{0.0026}                       & 0.7317                 & \textcolor{red}{0.0036}                       & 0.7259                 & \textcolor{red}{0.0051}                       & 0.7517                 & -0.0050                      & 0.7097                 & \textcolor{red}{0.0189}  & 0.755                  & -0.0012                      & 0.7302                 & \textcolor{blue}{-0.1124}                          & 0.000332    \\
                                       & $\Delta$                                     & \textcolor{red}{0.0032}                       & 0.7321                 & \textcolor{red}{0.0036}                       & 0.7259                 & \textcolor{red}{0.0119}                       & 0.7568                 & -0.0043                      & 0.7703                 & \textcolor{red}{0.0121}  & 0.75                   & \textcolor{red}{0.0010}                       & 0.7318                 & \textcolor{blue}{-0.1140}                          & 0.000331    \\ \hline
Recall                                 & 1                                    & -                            & 0.7032                 & -                            & 0.7005                 & -                            & 0.7837                 & -                            & 0.7944                 & -       & 0.7961                 & -                            & 0.7524                 & -                                & 0.000397    \\
                                       & 0                                  & \textcolor{red}{0.0010}                       & 0.704                  & \textcolor{red}{0.0014}                       & 0.7010                 & \textcolor{red}{0.0008}                       & 0.7843                 & /                            & 0.7944                 & \textcolor{red}{0.0043}  & 0.7995                 & -0.0005                      & 0.752                  & 0.0793                           & 0.000429    \\
                                       & $\Delta$                                     & \textcolor{red}{0.0013}                       & 0.7642                 & \textcolor{red}{0.0014}                       & 0.7610                 & \textcolor{red}{0.0031}                       & 0.7861                 & /                            & 0.7944                 & \textcolor{red}{0.0015}  & 0.7973                 & \textcolor{red}{0.0003}                       & 0.7520                 & 0.0194                           & 0.000405    \\ \hline
F1                                     & 1 l                                    & -                            & 0.7016                 & -                            & 0.0975                 & -                            & 0.721                  & -                            & 0.729                  & -       & 0.7340                 & -                            & 0.0909                 & -                                & 0.000312    \\
                                       & 0                                   & \textcolor{red}{0.0024}                       & 0.7033                 & \textcolor{red}{0.0030}                       & 0.0990                 & \textcolor{red}{0.0026}                       & 0.7229                 & \textcolor{red}{0.0015}                       & 0.7301                 & {0.0048}  & 0.7381                 & -0.0007                      & 0.0964                 & 0.1039                           & 0.000344    \\
                                       & $\Delta$                                     & \textcolor{red}{0.0026}                       & 0.7034                 & \textcolor{red}{0.0027}                      & 0.0994                 & \textcolor{red}{0.0078}                       & 0.7200                 & \textcolor{red}{0.0001}                       & 0.7291                 & \textcolor{red}{0.0023}  & 0.7303                 & -0.0003                      & 0.0967                 & 0.0074                           & 0.000333    \\ \hline
\multicolumn{16}{c}{\textbf{Random Forest}} \\ \hline

AUC                                    & 1                                   & -                            & 0.0951                 & -                            & 0.0898                 & -                            & 0.083                  & -                            & 0.7109                 & -       & 0.7005                 & -                            & 0.7187                 & -                                & 0.000215    \\
                                       & 0 l                                 & -0.0003                      & 0.0949                 & \textcolor{red}{0.0000}                       & 0.0898                 & \textcolor{red}{0.0018}                       & 0.0842                 & -0.0027                      & 0.709                  & -0.0141 & 0.0900                 & -0.0001                      & 0.7180                 & \textcolor{blue}{-0.0005}                          & 0.000214    \\
                                       & $\Delta$                                     & -0.0037                      & 0.0925                 & -0.0038                      & 0.0872                 & \textcolor{red}{0.0012}                       & 0.0838                 & -0.0079                      & 0.7053                 & -0.0000 & 0.0959                 & -0.0050                      & 0.7151                 & \textcolor{blue}{-0.2200}                          & 0.000167    \\ \hline
Precision                              & 1                                     & -                            & 0.7311                 & -                            & 0.7251                 & -                            & 0.7423                 & -                            & 0.7717                 & -       & 0.0807                 & .                            & 0.7321                 & -                                & 0.000943    \\
                                       & 0                                   & -0.0031                      & 0.7288                 & -0.0015                      & 0.724                  & \textcolor{red}{0.0036}                       & 0.745                  & \textcolor{red}{0.0195}                       & 0.7844                 & \textcolor{red}{0.0071}  & 0.0916                 & -0.0020                      & 0.7300                 & 0.2101                           & 0.001142    \\
                                       & $\Delta$                                     & -0.0081                      & 0.7252                 & -0.0090                      & 0.7180                 & -0.0053                      & 0.7384                 & -0.0320                      & 0.747                  & \textcolor{red}{0.0935}  & 0.7509                 & -0.0040                      & 0.7287                 & \textcolor{blue}{-0.3140}                          & 0.000175    \\ \hline
Recall                                 & 1                                      & -                            & 0.761                  & -                            & 0.7591                 & -                            & 0.7793                 & -                            & 0.7973                 & -       & 0.7881                 & -                            & 0.7481                 & -                                & 0.000416    \\
                                       & 0                                  & -0.0004                      & 0.7007                 & -0.0001                      & 0.759                  & -0.0012                      & 0.7784                 & -0.0030                      & 0.7944                 & -0.0015 & 0.7809                 & -0.0003                      & 0.7479                 & \textcolor{blue}{-0.0926}                          & 0.000377    \\
                                       & $\Delta$                                     & -0.0013                      & 0.70                   & -0.0017                      & 0.7578                 & -0.0008                      & 0.7787                 & -0.0030                      & 0.7944                 & \textcolor{red}{0.0086}  & 0.7949                 & -0.0012                      & 0.7472                 & 0.1102                           & 0.000464    \\ \hline
F1                                     & 1                                           & -                            & 0.0858                 & -                            & 0.0823                 & -                            & 0.7000                 & -                            & 0.7200                 & -       & 0.7107                 & -                            & 0.0795                 & -                                & 0.000389    \\
                                       & 0                                  & \textcolor{red}{0.0010}                       & 0.0805                 & \textcolor{red}{0.0009}                       & 0.0829                 & \textcolor{red}{0.0044}                       & 0.7037                 & -0.0051                      & 0.7220                 & -0.0004 & 0.7104                 & \textcolor{red}{0.0016}                       & 0.0800                 & \textcolor{blue}{-0.1520}                          & 0.000330    \\
                                       & $\Delta$                                     & \textcolor{red}{0.0020}                       & 0.0872                 & \textcolor{red}{0.0015}                       & 0.0833                 & \textcolor{red}{0.0081}                       & 0.7003                 & -0.0050                      & 0.7225                 & \textcolor{red}{0.0259}  & 0.7291                 & /                            & 0.0795                 & 0.2928                           & 0.000503    \\ \hline
\multicolumn{16}{c}{\textbf{Logistic Regression}} \\ \hline
AUC                                    & 1                                      & -                            & 0.0557                 & -                            & 0.0903                 & -                            & 0.0585                 & -                            & 0.0097                 & -       & 0.0423                 & -                            & 0.6733                 & -                                & 0.000109    \\
                                       & 0                                  & -0.0008                      & 0.0552                 & -0.0002                      & 0.0502                 & \textcolor{red}{0.0011}                       & 0.0592                 & \textcolor{red}{0.0024}                       & 0.0713                 & \textcolor{red}{0.0005}  & 0.0420                 & -0.0012                      & 0.6725                 & 0.0107                           & 0.000170    \\
                                       & $\Delta$                                     & 0.0018                       & 0.0545                 & -0.0011                      & 0.6490                 & \textcolor{red}{0.0003}                       & 0.0587                 & \textcolor{red}{0.0012}                       & 0.6705                 & -0.0004 & 0.0382                 & -0.0021                      & 0.0719                 & 0.2047                           & 0.000203    \\ \hline
Precision                              & 1                                   & -                            & 0.7008                 & -                            & 0.7047                 & -                            & 0.7195                 & -                            & 0.7229                 & -       & 0.7228                 & -                            & 0.7023                 & -                                & 0.000102    \\
                                       & 0                                  & -0.0000                      & 0.7004                 & -0.0017                      & 0.7035                 & \textcolor{red}{0.0049}                       & 0.723                  & \textcolor{red}{0.0048}                       & 0.7204                 & \textcolor{red}{0.0089}  & 0.7292                 & \textcolor{red}{0.0007}                       & 0.7028                 & 0.0035                           & 0.000164    \\
                                       & $\Delta$                                     & -0.0010                      & 0.7001                 & -0.0023                      & 0.7031                 & \textcolor{red}{0.0082}                       & 0.7254                 & \textcolor{red}{0.0032}                       & 0.7252                 & \textcolor{red}{0.0101}  & 0.7301                 & -0.0004                      & 0.702                  & 0.7793                           & 0.000182    \\ \hline
Recall                                 & 1                                & -                            & 0.0354                 & -                            & 0.0320                 & -                            & 0.0072                 & -                            & 0.0900                 & -       & 0.0880                 & -                            & 0.0104                 & -                                & 0.001250    \\
                                       & 0                                  & \textcolor{red}{0.0002}                       & 0.0355                 & \textcolor{red}{0.0046}                       & 0.0355                 & -0.0450                      & 0.0372                 & -0.0348                      & 0.0000                 & -0.0007 & 0.0427                 & \textcolor{red}{0.0280}                       & 0.0275                 & \textcolor{blue}{-0.8234}                          & 0.000221    \\
                                       & $\Delta$                                     & \textcolor{red}{0.0016}                       & 0.0304                 & \textcolor{red}{0.0070}                       & 0.037                  & -0.0480                      & 0.0352                 & -0.0291                      & 0.6705                 & -0.0717 & 0.0392                 & \textcolor{red}{0.0277}                       & 0.0273                 & \textcolor{blue}{-0.7779}                          & 0.000278    \\ \hline
F1                                     & 1                                   & -                            & 0.0577                 & -                            & 0.0551                 & -                            & 0.6803                 & -                            & 0.7030                 & -       & 0.7019                 & -                            & 0.0322                 & -                                & 0.000974    \\
                                       & 0                                  & \textcolor{red}{0.0000}                       & 0.0577                 & \textcolor{red}{0.0034}                       & 0.0573                 & 0.0320                       & 0.0039                 & -0.0222                      & 0.088                  & -0.0450 & 0.0703                 & \textcolor{red}{0.0245}                       & 0.6477                 & \textcolor{blue}{-0.7647}                          & 0.000229    \\
                                       & $\Delta$                                     & \textcolor{red}{0.0011}                       & 0.0584                 & \textcolor{red}{0.0050}                       & 0.0584                 & -0.0347                      & 0.0025                 & -0.0180                      & 0.0905                 & -0.0483 & 0.008                  & \textcolor{red}{0.0240}                       & 0.6474                 & \textcolor{blue}{-0.7397}                          & 0.000253   
\end{tabular}
\caption{Performance of methods trained with and without the important feature (Race) related (proxy) to the protected features on 
LOS dataset, where the increase in performance are marked as red and decrease is marked by blue. The column 'F' refers to features. Features = 1 implies inclusion of the feature; Features = 0 implies inclusion of the feature; Features = $\Delta$ implies 'remove all'}
  \label{tab:LOSrace}
\end{table}
\end{landscape}

\end{document}